\documentclass[runningheads]{llncs}
\usepackage{graphicx}

\usepackage{tikz}
\usepackage{comment}
\usepackage{amsmath,amssymb} 
\usepackage{color}

\usepackage[accsupp]{axessibility}  
\usepackage{graphics}
\usepackage{booktabs}
\usepackage{bbding}
\usepackage{multirow}
\usepackage{colortbl}
\usepackage{subfig}
\usepackage[misc]{ifsym}
\usepackage[colorlinks,linkcolor=red]{hyperref}

\def \algname{\texttt{\textbf{UniHead}}}
\def \etc{\emph{etc.}}

\begin{document}
\pagestyle{headings}
\mainmatter
\def\ECCVSubNumber{2746}  

\title{Unifying Visual Perception by Dispersible Points Learning} 

\titlerunning{Unifying Visual Perception by Dispersible Points Learning}
\author{Jianming Liang\inst{1,2}\textsuperscript{*} \and
Guanglu Song\inst{2} \and
Biao Leng \inst{1} \and
Yu Liu\inst{2}\textsuperscript{\dag}}
\authorrunning{J. Liang et al.}
\institute{School of Computer Science and Engineering, Beihang University \\ \and
SenseTime Research \\
\email{ljmmm1997@gmail.com},
\email{songguanglu@sensetime.com}, \\
\email{lengbiao@buaa.edu.cn},
\email{liuyuisanai@gmail.com}
}
\footnotetext{\textsuperscript{*} Work is done during the internship at SenseTime.}
\footnotetext{\textsuperscript{\dag} Corresponding author}
\maketitle

\begin{abstract}
We present a conceptually simple, flexible, and universal visual perception head for variant visual tasks, e.g., classification, object detection, instance segmentation and pose estimation, and different frameworks, such as one-stage or two-stage pipelines.
Our approach effectively identifies an object in an image while simultaneously generating a high-quality bounding box or contour-based segmentation mask or set of keypoints.
The method, called \texttt{\textbf{UniHead}}, views different visual perception tasks as the dispersible points learning via the transformer encoder architecture.
Given a fixed spatial coordinate, \algname{} adaptively scatters it to different spatial points and reasons about their relations by transformer encoder.
It directly outputs the final set of predictions in the form of multiple points, allowing us to perform different visual tasks in different frameworks with the same head design.
We show extensive evaluations on ImageNet classification and all three tracks of the COCO suite of challenges, including object detection, instance segmentation and pose estimation.
Without bells and whistles, \algname{} can unify these visual tasks via a single visual head design and achieve comparable performance compared to expert models developed for each task.
We hope our simple and universal \algname{} will serve as a solid baseline and help promote universal visual perception research.
Code and models are available at \url{https://github.com/Sense-X/UniHead}.
\end{abstract}

\section{Introduction}
Image classification~\cite{ImageNet}, object detection~\cite{VOC,COCO}, instance segmentation~\cite{COCO,Cityscapes} and human pose estimation~\cite{COCO,MPII} are the vital visual perception tasks in computer vision.
The vision community has rapidly improved results by developing robust feature representation.
Regardless of the development of the powerful backbone, in large part, these advances are inseparable from the task-aware visual head structure design, such as TSD~\cite{TSD}, CondInst~\cite{CondInst} and CPN~\cite{CPN}, or the elaborate frameworks construction, \emph{e.g.}, one-stage detectors~\cite{RetinaNet,SSD,FCOS} and two-stage detectors~\cite{FasterRCNN,CascadeRCNN}.
These methods are conceptually experienced and introduce task exclusivity, \emph{e.g.}, TSD~\cite{TSD} developed in object detection cannot be migrated to pose estimation.
Our goal in this work is to develop a comparably generalized feature representation learning with task-agnostic structure design for \emph{unifying visual perception}.

\begin{figure}[t]
  \centering
  \subfloat[Prediction targets in different visual tasks.]{
    \begin{minipage}[t]{0.7\textwidth}
    \centering
    \includegraphics[width=\textwidth]{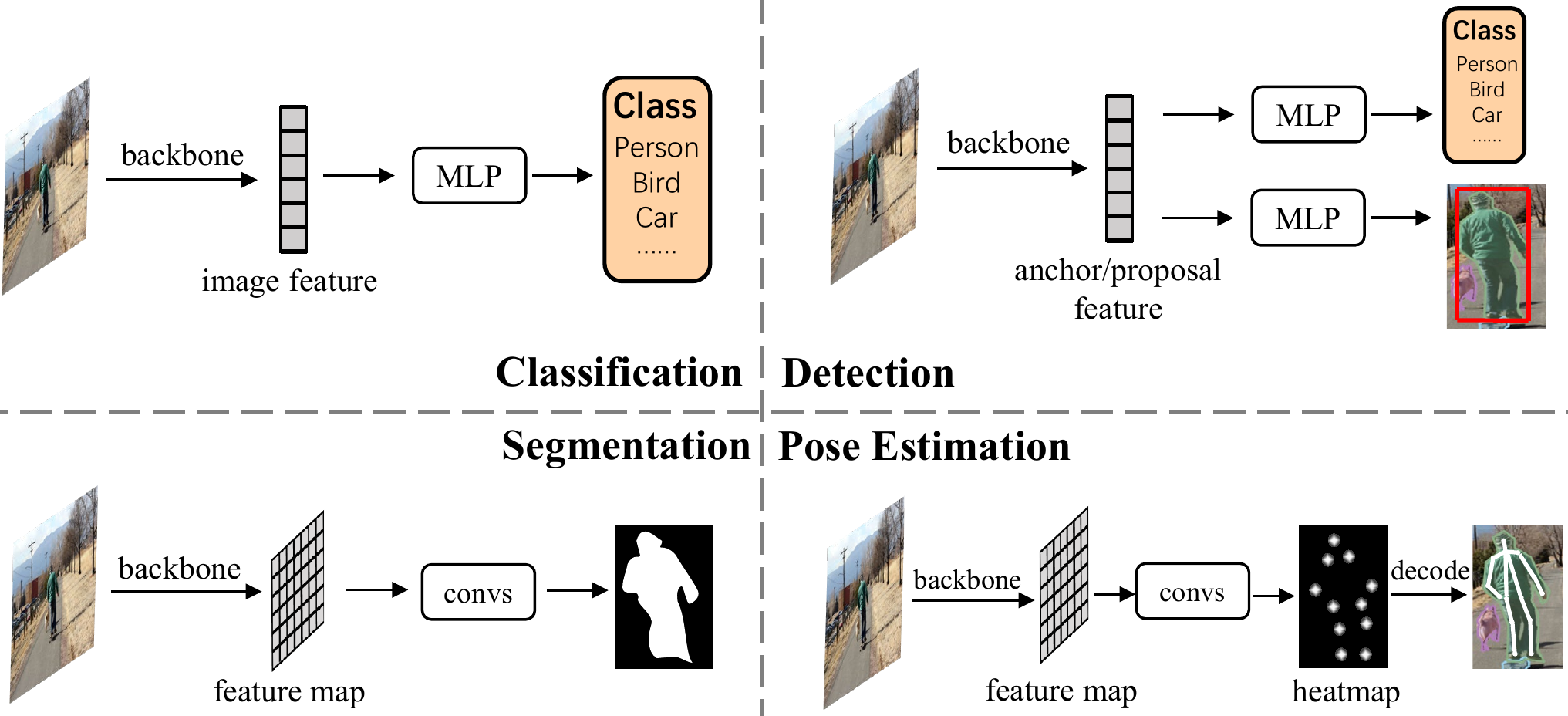}
    \end{minipage}
  } \\
  \subfloat[Unifying visual perception by \algname{}.]{
    \begin{minipage}[t]{0.7\textwidth}
    \centering
    \includegraphics[width=\textwidth]{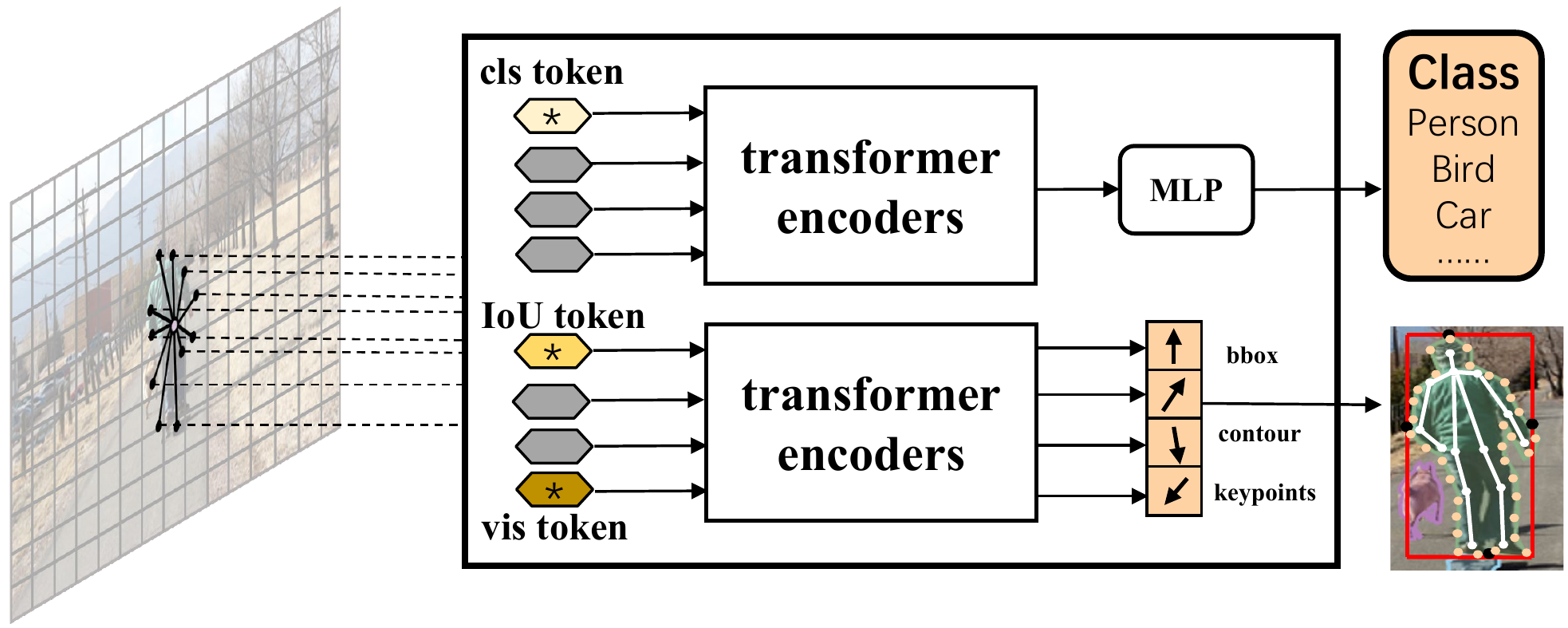}
    \end{minipage}
  }
  \caption{(a). Illustration of the typical pipelines for different visual tasks. Different sub-tasks require different prediction targets and different feature structures. (b). Illustration of the \algname{} design. Given a fixed spatial coordinate, \algname{} adaptively scatters it to different spatial points and reasons about their relations by transformer encoders. It directly outputs a set of predictions in the form of multiple points to perform different visual tasks.}
  \label{fig:task}
\end{figure}

The main barriers behind this are: 1) As shown in Fig.\ref{fig:task}(a), the different prediction targets  
force the visual perception into different sub-tasks, \emph{i.e.}, \emph{a class label} for image classification, \emph{a bounding box} for object detection, \emph{a pixel-wised mask} for instance segmentation, and \emph{a group of landmarks} for pose estimation. 2) How to conduct a task-agnostic head module which can generalize to all sub-tasks and frameworks while achieving good results?
Given this, one might expect a complex head design is required to solve these barriers. 
However, we show that a surprisingly simple, flexible, and universal head module can easily generalize to different visual tasks or frameworks and surpass prior expert models in each individual task.

Our method, called \algname{}, can be directly migrated to variant visual frameworks, \emph{e.g.}, Faster RCNN~\cite{FasterRCNN}, FCOS~\cite{FCOS} and ATSS~\cite{ATSS}, by formulating the prediction targets as the dispersible points learning.
As shown in Fig.\ref{fig:task}(b), \algname{} is built upon any network backbone and the prediction targets for different tasks can be achieved by a basic yet effective points estimation.
Given a fixed spatial coordinate, \algname{} adaptively scatters it to different spatial points and reasons about their relations by several stacked transformer encoders.
It directly outputs the final set of predictions in the form of multiple points, which is robust to geometric variations an object can exhibit, including scale, deformation, and orientation.
For \emph{image classification}, the points directly predict the object class.
For \emph{object detection}, the points are placed along the four edges of a bounding box.
For \emph{instance segmentation}, the points are evenly distributed along the instance mask contour.
For \emph{pose estimation}, the position of points conforms to the pose distribution of the training data.

Furthermore, we found it essential to adapt the initial position of the points according to different prediction targets.
This can effectively alleviate the difficulty of optimization under the requirement of fitting objects with different scales and orientations.
Additionally, the \algname{} only adds a small computational overhead, enabling a universal system and rapid experimentation.

Without bells and whistles, \algname{} can be equipped with popular backbones on different visual tasks, such as ResNet~\cite{ResNet}, ResNeXt~\cite{ResNeXt}, Swin Transformer~\cite{SwinTransformer}, etc.
It excels on the ImageNet~\cite{ImageNet} classification and all three tracks of the COCO~\cite{COCO} suite of challenges, including object detection, instance segmentation, and human pose estimation. 
We conduct extensive experiments to showcase the generality of our \algname{}.
By viewing each task as the dispersible points learning via the transformer encoder architecture, \algname{} can perform comparably without any special design for specific tasks.
\algname{}, therefore, can be seen more broadly as a universal head module for visual perception and easily migrated to more complex tasks.

To summarize, our contributions are as follows:

1) We develop a comparably generalized dispersible points learning method for unifying visual perception. We hope our work can inspire the vision community to explore a unified vision framework.

2) We introduce the transformer encoder to reason about the relations of dispersible points and the adaptively points initialization to handle the geometric variations
an object can exhibit, including scale, deformation, and orientation. 

3) Detailed experiments on ImageNet~\cite{ImageNet} and MS-COCO~\cite{COCO} datasets show that \algname{} can easily generalize to different tasks while obtaining comparable performance compared to the expert models developed in individual tasks.

\section{Related Work}
Image classification~\cite{ImageNet}, object detection~\cite{VOC,COCO}, instance segmentation~\cite{COCO,Cityscapes} and pose estimation~\cite{MPII,COCO} are four popular tasks in computer vision. 
They all benefit a lot from the development of deep neural networks~\cite{ResNet,HRNet}. 
Among them, image classification~\cite{AlexNet} was the first to be applied with CNNs. 
The performance was improved by a considerable margin. After that, researchers are devoted to designing powerful backbones~\cite{ResNet,ResNeXt,DenseNet}, which also give lift to other instance-level tasks, such as object detection~\cite{FasterRCNN,RetinaNet} and human pose estimation~\cite{HRNet}.

For object detection, it requires bounding box level location and category information of interested instances in an image. 
The methods can be roughly categorized into three types: \textbf{Two-stage}, \textbf{One-stage} and \textbf{DETR} detectors. 
\textbf{Two-stage} methods detect a series of region proposals at first and refine them in the second stage. Faster RCNN~\cite{FasterRCNN} is a popular pipeline of the two-stage method, which also includes R-FCN~\cite{R-FCN}, Cascade RCNN~\cite{CascadeRCNN}, Grid RCNN~\cite{GridRCNN}, \etc. 
\textbf{One-stage} methods predict locations and class scores on a large amount of pre-defined spatial candidates. 
They can be further divided into two types: anchor-based and anchor-free detectors. 
Anchor-based methods use anchor boxes as an initial set, such as SSD~\cite{SSD} and RetinaNet~\cite{RetinaNet}. For anchor-free methods, some methods make dense predictions on spatial points, such as CenterNet (objects as points)~\cite{CenterNetObj}, FCOS~\cite{FCOS} and RepPoints~\cite{RepPoints}. 
And some other works obtain a keypoint heatmap first and get objects by grouping them. 
CornerNet~\cite{CornerNet}, ExtremeNet~\cite{ExtremeNet} and CenterNet (keypoint triplets)~\cite{CenterNetKeyp} fall into this category.
\textbf{DETR} methods, such as DETR~\cite{DETR}, Deformable DETR~\cite{DeformableDETR} and Conditional DETR~\cite{ConditionalDETR}, propose to detect objects by decoding a pre-defined set of object queries with transformers. 
These queries are optimized one-to-one with ground truths so there is no need for NMS as post-processing. 
Such a way of one-to-one label assignment also inspires other works like Sparse RCNN~\cite{SparseRCNN}.

For instance segmentation, it requires mask and class information for instances. 
The methods can be categorized into two types: \textbf{mask-based} and \textbf{contour-based}. \textbf{Mask-based} methods predict binary mask directly, which can further be divided into local-mask and global-mask methods. 
Most local-mask methods include two stages: the first one for instance detection and the second one for instance mask generation, such as Mask RCNN~\cite{MaskRCNN}, PANet~\cite{PANet} and PointRend~\cite{PointRend}. Global-mask methods usually predict the mask for the whole image and leverage dynamic mask filters to decode masks for different instances, such as YOLACT~\cite{Yolact} and CondInst~\cite{CondInst}. \textbf{Contour-based} methods obtain instance masks by predicting object boundaries. 
PolarMask~\cite{PolarMask} and DeepSnake~\cite{DeepSnake} are two typical works using this idea.

For human pose estimation, it requires the keypoint locations (\emph{e.g.} nose, eyes, knees) for multiple humans in an image. There are mainly two kinds of approaches: \textbf{heatmap-based} and \textbf{regression-based}. 
\textbf{Heatmap-based} methods use a multi-class classifier to generate keypoint heatmaps and compose them with clustering and grouping procedures, such as CPN~\cite{CPN}, HRNet~\cite{HRNet} and DARK~\cite{DARK}. \textbf{Regression-based} methods, including Integral~\cite{Integral} and CenterNet~\cite{CenterNetObj}, \etc, predict coordinates of keypoints directly. 
It is more simple to plug them into existing end-to-end learning frameworks.

Mask R-CNN~\cite{MaskRCNN}, PointSetNet~\cite{PointSet} and LSNet~\cite{LSNet} achieved merging object detection, instance segmentation and pose estimation into one network.
Besides these tasks, \algname{} can be extended to image classification.
Furthermore, \algname{} can also be simply embedded in variant types of architectures, \emph{e.g.}, anchor-free, anchor-based, and two-stage detectors, showing powerful ability on task and framework generalization.

\begin{figure}[t]
  \centering
  \includegraphics[width=\linewidth]{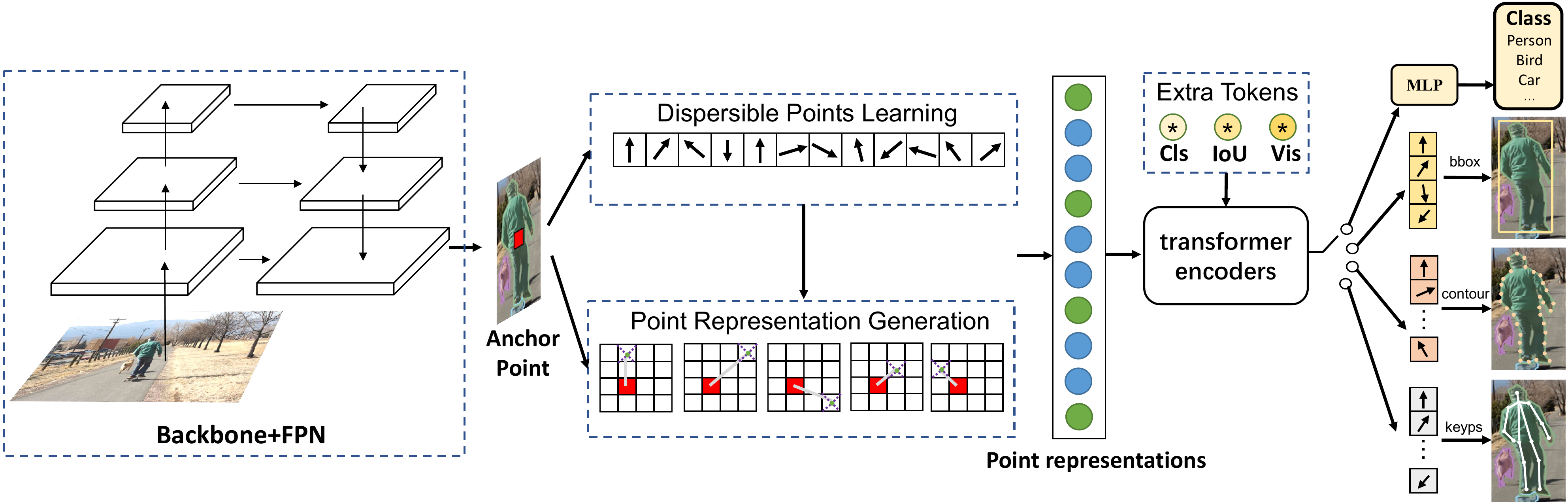}
  \caption{A typical pipeline of \algname{}. At first, most methods of location-sensitive tasks contain a backbone and the feature pyramid (not used in the image classification task) to extract feature maps. Then, for an anchor point, \algname{} obtains multiple points via dispersible points learning. To generate point representations, bilinear interpolation is performed on the feature map according to point coordinates, which is denoted in dotted line. The obtained features will be concatenated with extra learnable tokens if necessary, and sent to corresponding transformer encoders to complete variant visual tasks.}
  \label{fig:method}
\end{figure}

\section{Method}
In this paper, we introduce the \algname{}, a generalized visual head. 
It can be applied to different detection frameworks, such as Faster RCNN~\cite{FasterRCNN}, FCOS~\cite{FCOS} and ATSS~\cite{ATSS}, as well as different tasks including classification, object detection, instance segmentation and pose estimation.
In this section, we first describe the design principle of \algname{} and then detail the adaptation to different visual tasks and different visual frameworks. Finally, we delve into the inherent advantage of \algname{} over other methods.

\subsection{UniHead}

In \algname{}, given a fixed spatial coordinate ($\mathcal{A}_x, \mathcal{A}_y$) (referred as \textbf{anchor point}), \emph{i.e.}, \emph{center point of a proposal or a point in the feature map}, it adaptively scatters it to different spatial points and reasons about the relations of them by several stacked transformer encoders.
As shown in Fig.\ref{fig:method}, 
\algname{} adopts the sequentially three-stage procedure to seek for the scattered point representations.
In the first stage, it will generate the anchor representation $\mathcal{F}_{x,y}$ according to the anchor coordinate or region proposal.
For one-stage or anchor-free detectors, it is designated by the feature representation in the corresponding coordinate of the feature map.
For the two-stage detectors, the feature generated by RoI Pooling~\cite{FasterRCNN} is used.
In the second stage, $K$ scattered points are generated by:
\begin{equation}
\begin{split}
    P_{x_i} &= \mathcal{A}_x + s_x \cdot \Delta x_i\\
    P_{y_i} &= \mathcal{A}_y + s_y \cdot \Delta y_i, 
\end{split}
\label{eq1}
\end{equation}
where $(\Delta x_i, \Delta y_i) = f(\mathcal{F}_{x,y}; w_i)$. 
$f$ is a simple multi-layer perceptron and $w_i$ is the learnable parameter.
$(s_x, s_y)$ is the computed scalar to modulate the magnitude of the $(\Delta x_i, \Delta y_i)$.
Specifically, $(s_x, s_y)$ is the width and height of the region proposal in a two-stage detector, the anchor scale in a one-stage detector, and the model stride in an anchor-free detector.
In the final stage, instead of quantizing a floating-number of $(P_{x_i}, P_{y_i})$, we perform bilinear interpolation to generate the point representations $\mathcal{F}_{x_i,y_i}, i\in [1,K]$.

To better reason about the relations of these scattered point representations and generate more informative features, we introduce the transformer operator to capture the correlative dependence between them.
To improve the robustness of different visual tasks, we insert a task-aware token embedding by:
\begin{equation}
\begin{split}
    z_0 = [{\bf T_{task}}; \mathcal{F}_{x_1,y_1}; \mathcal{F}_{x_2,y_2}; \dots; \mathcal{F}_{x_K,y_K}],  
    \label{eq2}
\end{split}
\end{equation}
where $\bf{T_{task}}$ can be $\bf{T_{class}}$, $\bf{T_{IoU}}$, and $\bf{T_{visibility}}$ for image classification, object detection and pose estimation, respectively.
The computation in transformer encoders for point representations can be formulated as:
\begin{equation}
\begin{split}
    &z^{'}_{l} = {\rm MHSA}({\rm LN}(z_{l-1})) + z_{l-1}, \qquad l = 1\dots L, \\
    &z_l = {\rm MLP}({\rm LN}(z^{'}_{l})) + z^{'}_{l}, \qquad l = 1\dots L, \\
    &[{\bf T^{'}_{task}}; \mathcal{F}^{'}_{x_1,y_1}; \mathcal{F}^{'}_{x_2,y_2}; \dots; \mathcal{F}^{'}_{x_K,y_K}]=z_{L},  \label{eq3}
\end{split}
\end{equation}
where ${\rm MHSA}$ means multi-head self attention in~\cite{Transformer}, $\rm LN$ indicates layer normalization~\cite{LN}, $\rm MLP$ is a multi-layer perceptron.
Formally, during training, we use $L$ transformer encoders, and the final output $z_{L}$ will be adapted to different visual tasks to perform the task-aware prediction.

\subsection{Adaptation to Different Visual Tasks}
\noindent\textbf{Image Classification.} 
For image classification, we directly use the final feature map to perform dispersible points learning. The anchor point is set as the center of the input image and the corresponding scales are the input scale. 
We choose to align the classifier setting with standard vision transformers, \emph{i.e.}, only leveraging classification token instead of all tokens in the classifier. 
The training can be formulated as:
\begin{equation}
    \mathcal{L}_{\rm cls} = {\rm CrossEntropy}({\rm softmax}({\rm MLP}(\bf{T^{'}_{cls}})),y).
\end{equation}
In the above $\bf y$ specifies the ground-truth class and
$\rm{MLP}$ is a single fully-connected layer predicting the model’s probability for the class with label $\bf y$.

\noindent\textbf{Object Detection.}
\algname{} can be applied to a variety of detectors, such as Faster R-CNN~\cite{FasterRCNN}, FCOS~\cite{FCOS}, \etc, without changing the backbone network structure, and the manner of label assignment.
Specially, we concatenate a learnable token $\bf{T_{IoU}}$ as a replacement for the IoU branch. 
After passing through all transformer blocks, the $\bf{T^{'}_{IoU}}$ is used to predict IoU, which will be multiplied by class prediction to get final scores at inference time. 
The $\mathcal{F}^{'}_{x_i,y_i}$ is used to predict the offset for point $(P_{x_i}, P_{y_i})$.
There are:
\begin{equation}
    (P^{'}_{x_i}, P^{'}_{y_i}) = (P_{x_i}, P_{y_i}) + {\rm MLP}(\mathcal{F}^{'}_{x_i,y_i})
    \odot (s_x, s_y)
    , \label{eq4}
\end{equation}
where $\odot$ denotes element-wise multiplication, and the $\rm{MLP}$ is a single fully-connected layer shared between different points.
The predicted bounding box can be computed by $B^{'}$ = $({\rm min}\{P^{'}_{x_i}\}, {\rm min}\{P^{'}_{y_i}\}, {\rm max}\{P^{'}_{x_i}\}, {\rm max}\{P^{'}_{y_i}\})$, $i \in [1,K]$.

For the classification branch, it performs the same computational manner as \algname{} in image classification.
For regression, it shares $z_0$ with the classification branch to reduce the computational cost of point representation generation.
Our loss function for detection is defined as:
\begin{equation}
    \mathcal{L}_{loc} = -\frac{1}{n}\sum_{j=1}^n L_1(B^{'}_j, B_j),
\end{equation}
where $j$ is the index of positive samples, $B^{'}_j$ is the predicted box and $B_j$ is the ground truth.
Other kinds of detection loss can also be used, \emph{e.g.}, GIoU loss~\cite{GIoU}.

\noindent\textbf{Instance Segmentation.}
For instance segmentation, we view this task as the contour-based regression.
\algname{} is placed at the output of the backbone to generate the points $P^{'}_{x_i,y_i}$ by Eq.\ref{eq1}, Eq.\ref{eq2}, Eq.\ref{eq3} and Eq.\ref{eq4}.
To align the point number between scattered points and the contour points in training data, we uniformly add new points, or delete points with the shortest edge until the target number is met, which is similar to Deep Snake~\cite{DeepSnake}. 
All ground truth points are clockwise arranged around the contour line. 
The scattered points $\{P^{'}_{x_i,y_i}, i\in [1,K]\}$ are uniformly and clockwisely perform one-to-one matching with them.

Besides, some objects are split into several components due to occlusions. To overcome this problem, we simply follow PolarMask~\cite{PolarMask} and directly treat them as multiple objects. During training, we use $L_1$ loss to optimize each point:
\begin{equation}
    \label{eq:segm}
    \mathcal{L}_{seg} = \frac{1}{n}\sum_{i=1}^n L_1(P^{'}_{x_i,y_i}, P_{x_i,y_i}),
\end{equation}
where $P^{'}_{x_i,y_i}$ is the predicted point and $P_{x_i,y_i}$ is the corresponding ground truth.

\noindent\textbf{Pose Estimation}
The overall design of pose estimation is consistent with instance segmentation, except that an extra token $\bf{T_{visibility}}$ is introduced to predict the visibility of keypoints. The number $K$ of predicted points is aligned with keypoint number in the dataset. For pose estimation, each keypoint has a clear definition, like nose, eyes, \etc, which makes it possible to build one-to-one connection with dispersible points. 
$l_1$ loss is adopted to train the keypoint localization branch, same as Eq.\ref{eq:segm}. For the training of keypoint visibility prediction, we use standard binary cross entropy loss.

\subsection{Adaptation to Different Visual Frameworks}
\noindent\textbf{Two-stage Framework.}
\algname{} is applied to region proposals in the two-stage framework. 
Each proposal is represented as a combination of an anchor point ($\mathcal{A}_x, \mathcal{A}_y$) and its scale ($s_x, s_y$).
The offsets $(\Delta x_i, \Delta y_i)$ are generated from the proposal feature extracted with RoI Pooling or RoI Align.
Without other modifications, \algname{} now can be directly used on a two-stage framework.

\noindent\textbf{One-stage Framework.}
\algname{} is applied on dense spatial points in the one-stage framework.
For anchor-free methods, ($\mathcal{A}_x, \mathcal{A}_y$) and ($s_x, s_y$) are a point and the stride of the feature map. 
For anchor-based methods, ($\mathcal{A}_x, \mathcal{A}_y$) and ($s_x, s_y$) are the center point and the scale of an anchor.
The offsets $(\Delta x_i, \Delta y_i)$ are generated using a 1$\times$1 convolutional layer.

\begin{figure}[t]
  \centering
  \includegraphics[width=0.7\linewidth]{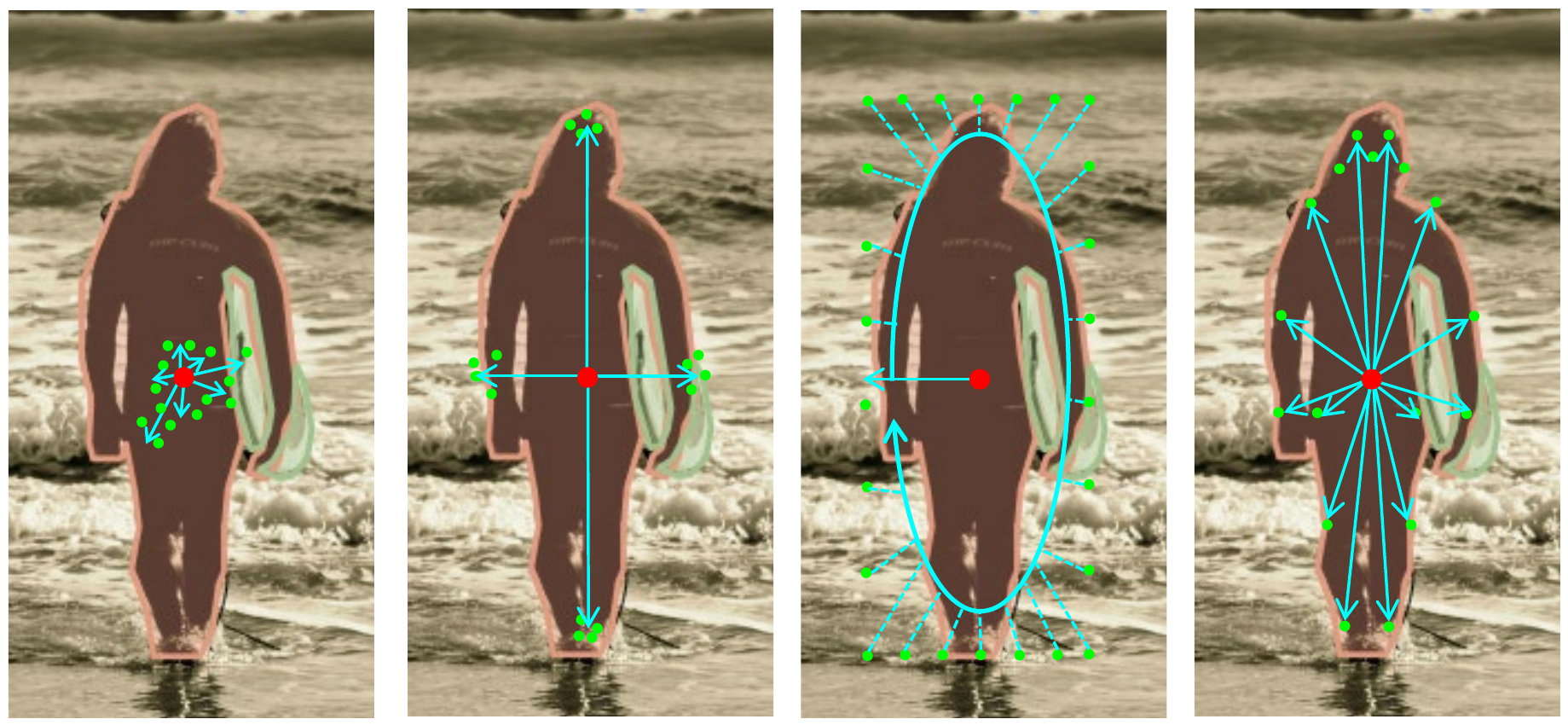}
  \caption{Ways of point initialization for different tasks. From left to right: image classification, object detection, instance segmentation, pose estimation.}
  \label{fig:initialize}
\end{figure}

\subsection{UniHead Initialization}
To effectively alleviate the difficulty of optimization under the requirement of fitting objects with different scales and orientations, the result points are initialized in a more appropriate way for different tasks, which is illustrated in Fig.\ref{fig:initialize}. 
For image classification, points are casually scattered around the anchor point.
For object detection, points are divided into four groups placed at the bottom, top, left, and right of the anchor point, respectively.
For instance segmentation, first we set a 2D reference vector that starts from the anchor point. 
Based on the direction of this vector, the points are uniformly and clockwise initialized on the edge of a pseudo box generated from the anchor point and its spatial scale.
For pose estimation, we calculate the average positions of different keypoints in the training dataset and use them to initialize points.

The initial point position is controlled by 
tuning the \textit{bias} of the last fully-connected layer in $\rm{MLP}$ used for offsets generation. Taking object detection as an example, the \textit{bias} for points at left, right, top and bottom are set to $[-0.5, 0]$, $[0.5, 0]$, $[0, -0.5]$ and $[0, 0.5]$, respectively.

\section{Experiments}
For image classification, experiments are conducted on the ILSVRC-2012 ImageNet~\cite{ImageNet} dataset with 1K classes and 1.3M images. We use Top-1 accuracy as the metric in classification experiments.

We also conduct experiments with different backbones on the MS-COCO 2017~\cite{COCO} dataset, including object detection, instance segmentation, and human pose estimation tasks. 
For these tasks, training is performed on the \textit{train} set, over 57K images for human pose estimation, and over 118K images for object detection and instance segmentation. 
For experiments of ablation studies, evaluation is conducted on the \textit{val} set. 
We also report performance on the \textit{test-dev} set to compare with the state-of-art methods. 
The mean average precision (AP) is used as the measurement in COCO experiments.
But the definition of AP varies with tasks. 
For object detection and instance segmentation, AP is calculated under different IoU thresholds (bounding box IoU or mask IoU). 
For human pose estimation, AP is calculated with object keypoint similarity (OKS).

\begin{table}[t]
    \caption{Ablation study on extra blocks for image classification task.}
    \centering
    \scalebox{0.8}{
    \begin{tabular}{c|c|c}
    \hline
    Method & GFLOPs & Top-1 acc. \\
    \hline
    ResNet-50 & 3.8 & 78.5 \\
    ResNet-50+extra blocks & 4.2 & 79.0 \\
    ResNet-50+UniHead & 4.1 & \textbf{79.5} \\
    \midrule
    \hline
    Swin-T & 4.5 & 81.2 \\
    Swin-T+extra blocks & 5.2 & 81.7 \\
    Swin-T+UniHead & 4.7 & \textbf{81.8} \\
    \hline
    Swin-B & 15.4 & 83.5 \\
    Swin-B+extra blocks & 16.7 & 83.6 \\
    Swin-B+UniHead & 15.7 & \textbf{83.9} \\
    \hline
    \end{tabular}}
    \label{tab:extra_block}
\end{table}

\begin{table}[t]
\centering
\caption{Ablation study on $\bf{T_{task}}$. 'Det.' and 'Keyp.' mean detection and pose estimation, respectively.}
\scalebox{0.8}{
\begin{tabular}{c |c | c c c| c c c}
\hline
Task & w/ ${\bf T_{task}}$ & AP & AP$_{.5}$ & AP$_{.75}$ & AP$_s$ & AP$_m$ & AP$_l$ \\
\hline
\multirow{2}{*}{Det.} & x & 41.6 & \textbf{61.2} & \textbf{44.8} & 23.4 & 45.1 & 56.2 \\
& \checkmark & \textbf{41.8} & 60.6 & 44.7 & \textbf{23.8} & \textbf{45.1} & \textbf{56.7} \\
\hline
\multirow{2}{*}{Keyp.} & x & 50.4 & 78.8 & 53.9 & - & 44.9 & 58.5 \\
& \checkmark & \textbf{50.7} & \textbf{78.9} & \textbf{54.6} & - & \textbf{45.5} & \textbf{58.5} \\
\hline
\end{tabular}}
\label{tab:task_token}
\end{table}

\subsection{Implementation Details}
In the image classification task, all models are trained using AdamW optimizer~\cite{AdamW} with 1e-4 initial learning rate, 0.05 weight decay, $\beta_1=0.9$, $\beta_2=0.999$ and a batch size of 1024. We train classification models for 300 epochs and use consine annealing scheduler to decrease learning rate. Data augmentations in ~\cite{Deit} are also used, \emph{e.g.}, mix up, label smoothing, \etc.

For other three tasks, we use different backbones including ResNet~\cite{ResNet}, ResNeXt~\cite{ResNeXt} and Swin Transformer~\cite{SwinTransformer} with weights pretrained on ImageNet~\cite{ImageNet}. 
For object detection, we use our \algname{} on different detection pipelines and follow their original hyper-parameters. 
For instance segmentation and pose estimation, the same settings as Faster RCNN~\cite{FasterRCNN} are used. During training, we adopt AdamW~\cite{AdamW} as the optimizer, with 1e-4 initial learning rate, 0.05 weight decay, $\beta_1=0.9$ and $\beta_2=0.999$. 
In our $1\times$ setting, we train our model with mini-batch size 16 for 13 epochs and decrease the learning rate by a factor of 10 at epoch 9 and 12. Unless specified, the input scale of images is [800, 1333] and no data augmentations except horizontal flipping are used in training. 
The hyper-parameter of newly-added transformers keeps the same as \cite{ViT}.

\begin{table}[t]
\centering
\caption{Ablation study on UniHead bias initialization strategy.}
\scalebox{0.8}{
\begin{tabular}{c|c|c c c}
\hline
Task & UniHead Initialization? & AP & AP$_{.5}$ & AP$_{.75}$ \\ \hline
\multirow{2}{*}{Det.} & x & 40.9 & 60.7 & 43.7 \\
& \checkmark & \textbf{41.6} & \textbf{61.2} & \textbf{44.8} \\
\hline
\multirow{2}{*}{Segm.} & x & 29.7 & \textbf{53.5}  & 28.9 \\ 
& \checkmark & \textbf{30.4} & 53.2 & \textbf{30.1} \\ 
\hline
\multirow{2}{*}{Keyp.} & x & 57.0 & 81.9 & 62.4 \\
& \checkmark & \textbf{57.9} & \textbf{82.6} & \textbf{63.9} \\
\hline
\end{tabular}}
\label{tab:bias_ini}
\end{table}

\begin{table}[t]
\centering
\caption{Ablation study on \textbf{point number}. Point number 8, 16, 24, 32 are tried.}
\scalebox{0.8}{
\begin{tabular}{c| c c c | c c c}
\hline  
$K$ & AP & AP$_{.5}$ & AP$_{.75}$ & AP$_s$ & AP$_m$ & AP$_l$ \\
\hline
8 &  40.8 & 59.3 & 43.7 & 22.3 & 44.3 & 54.6 \\
16 & 41.8 & 60.6 & 44.7 & 23.8 & \textbf{45.1} & 56.7 \\
24 & \textbf{41.8} & \textbf{60.7} & \textbf{44.7} & \textbf{23.9} & 44.8 & 56.5\\
32 & 41.5 & 60.3 & 44.5 & 22.8 & 44.6 & \textbf{56.8}\\
\hline
\end{tabular}}
\label{tab:point_num}
\end{table}

\subsection{Ablation Studies}
In this section, we conduct extensive ablation studies on ImageNet and COCO \textit{val} set to validate the effectiveness of \algname{} on classification and localization tasks, respectively. 
Specially, for localization task, we choose object detection and all models are trained on Faster RCNN~\cite{FasterRCNN} baseline with AdamW optimizer~\cite{AdamW} and ResNet-50 backbone for fair comparison.
We find that AdamW can stably improve the performance by $\sim1$\% AP compared to SGD.

\noindent\textbf{Extra Blocks.}
We add extra blocks to the classification backbone networks to align their FLOPs with \algname{}. 
Specifically, we append two bottlenecks to ResNet-50 ([3,4,6,5] for four stages) and two transformer blocks to Swin-T ([2,2,6,4] for four stages), whose results are shown in Table~\ref{tab:extra_block}.
Though additional layers can boost the performance, \algname{} can achieve better performance with similar FLOPs.
Also, we conduct the same experiment on Swin-B. 
We can see that when the model becomes bigger with higher FLOPs, extra blocks can hardly bring improvement. 
But UniHead achieves a continual performance boost.
All these results prove that improvement brought by \algname{} does not only account for its transformer blocks.

\noindent\textbf{Task Token.}
We also explore the influence of $\bf{T_{IoU}}$ and $\bf{T_{visibility}}$ on object detection and pose estimation, respectively. 
As is shown in Table~\ref{tab:task_token}, the introduction of $\bf{T_{task}}$ brings a slight improvement on both tasks, proving the effectiveness of task tokens. 
It is worth noting that though visibility prediction is not used in pose estimation evaluation, $\bf T_{visibility}$ still has a positive impact on training.

\noindent\textbf{UniHead Initialization.}
We replace our task-specific bias initialization with zero initialization on different tasks. 
Main results are shown in Table~\ref{tab:bias_ini}. 
It proves that a proper initialization can help the unified architecture learn the knowledge of different tasks more quickly.

\noindent\textbf{Point Number.}
We evaluate the performance of different point numbers in \algname{}, which is shown in Table~\ref{tab:point_num}. It shows that our head can benefit from the increasing number of points. But more points may bring overfitting and more computational cost. So we choose to use $K=16$ in our implementations.

\begin{table}[t]
\centering
\caption{Ablation study on \textbf{block number}. $L_{cls}$ and $L_{loc}$ denote transformer encoder block number of classification and localization, respectively. \#params means parameters of the detection head. The training and inference time is measured on a 16GB V100 GPU.}
\scalebox{0.8}{
\begin{tabular}{c|c|c c| p{40pt}<{\centering} p{40pt}<{\centering} |c c c}
\hline  
$L_{cls}$ & $L_{loc}$ & \#params & GFLOPs & Training (s/iter) & Inference (ms/img) & AP & AP$_{.5}$ & AP$_{.75}$ \\
\hline
\multicolumn{2}{c|}{baseline} & 14.3M & 215 & 0.38 & 82 & 38.8 & 59.9 & 42.1 \\
\hline
1 & 1 & 11.9M & 227 & 0.39 & 85 & 41.6 & 60.3 & 44.3 \\
2 & 2 & 12.7M & 239 & 0.40  & 87  & 41.6 & 60.4 & 44.7 \\
3 & 3 & 13.5M & 251 & 0.40 & 90 & 41.7 & 60.5 & 44.7 \\
4 & 4 & 14.3M & 263 & 0.41 & 92 &\textbf{42.0} & \textbf{60.6} & \textbf{45.0} \\
\hline
\end{tabular}}
\label{tab:block_num}
\end{table}

\begin{table}[t]
\centering
\caption{Ablation study on \textbf{different modules}. IoU prediction is not used in this table. "HD", "MHSA" and "DPL" mean head disentanglement, multi-head self attention and dispersible points learning, respectively.}
\scalebox{0.8}{
\begin{tabular}{c c c|c c c|c c c}
\hline
HD & MHSA & DPL & AP & AP$_{.5}$ & AP$_{.75}$ & AP$_s$ & AP$_m$ & AP$_l$ \\
\hline
x & x & x & 38.8 & 59.9 & 42.1 & 22.1 & 41.9 & 51.9 \\
\checkmark & x & x & 39.3 & 60.0 & 42.5 & 22.0 & 42.9 & 52.6 \\
\checkmark & \checkmark & x & 39.9 & 60.5 & 43.4 & 22.4 & 43.2 & 53.4 \\
\checkmark & x & \checkmark & 40.7 & \textbf{61.6} & 44.4 & 23.1 & 43.4 & 55.1 \\
\checkmark & \checkmark & \checkmark & \textbf{41.6} & 61.2 & \textbf{44.8} & \textbf{23.4} & \textbf{45.1} & \textbf{56.2} \\
\hline
\end{tabular}}
\label{tab:components}
\end{table}

\noindent\textbf{Block Number.}
We also analyze the influence of the number of transformer encoder blocks. 
As is shown in Table~\ref{tab:block_num}, we compare the performances, head parameters, FLOPs, training time, and inference time with baseline under different block number settings. 
Our head can benefit slightly from the increase in block numbers. 
Considering computational costs and the head capacity, we finally use $L_{cls}=2$ and $L_{loc}=3$ in our implementations.

\noindent\textbf{Head Disentanglement.}
To show that our method does not only benefit from the separated task heads, a Faster RCNN with sibling heads is given in the second row of Table~\ref{tab:components}. 
We simply remove the shared fully connected layers in the RCNN head and replace them with separated ones. 
We can observe that the improvement brought by head disentanglement (0.5 AP) is actually limited.

\noindent\textbf{Dispersible Points Learning and Multi-head Self Attention.}
In order to demonstrate the effectiveness of dispersible points learning and multi-head self attention, we conduct experiments with different head designs and compare them with our head (without IoU prediction). 
First, we take the output of RoI Align~\cite{MaskRCNN} as tokens directly (49 in total), and process them with disentangled transformer encoders. 
The result is in the third row of Table~\ref{tab:components}. 
We can see that though more points are used, it still performs worse than DPL with $K=16$.

Then, we leverage deformable RoI Pooling~\cite{DCN} as another form of dispersible points learning. Specifically, multiple offsets are generated in the same way and applied to deformable RoI Pooling for feature extraction. The result is shown in the fourth row of Table~\ref{tab:components}. It indicates that the combination of dispersible points learning and multi-head attention is more effective to capture semantic information within an instance.

\subsection{Generalization Ability}

\begin{table}[t]
\centering
\caption{Results of \algname{} with variant detection pipelines.}
\scalebox{0.8}{
\begin{tabular}{c| c c c }
\hline  
Method & AP & AP$_{.5}$ & AP$_{.75}$ \\
\hline
Faster RCNN~\cite{FasterRCNN}&  38.8 & 59.9 & 42.1 \\
+UniHead& \textbf{41.8} & \textbf{60.6} & \textbf{44.7} \\
\hline
Cascade RCNN~\cite{CascadeRCNN}&  42.1 & 60.8 & 45.3 \\
+UniHead& \textbf{43.0} & \textbf{61.5} & \textbf{46.2} \\
\hline
ATSS (anchor-based)~\cite{ATSS} & 39.5 & 58.1 & 42.2 \\
+UniHead & \textbf{40.6} & \textbf{58.3} & \textbf{44.2} \\ 
\hline
FCOS (w/o imprv.)~\cite{FCOS} & 37.1 & 56.3 & 39.1 \\
+UniHead& \textbf{39.7} & \textbf{57.9} & \textbf{42.6} \\
\hline
Mask RCNN~\cite{MaskRCNN} &  35.2 & 56.8 & 37.5 \\
+UniHead (mask)&  \textbf{37.0} & \textbf{57.9} & \textbf{39.9} \\
\hline
\end{tabular}}
\label{tab:det_pipelines}
\end{table}

\textbf{Detection Pipeline Generalization.}
We evaluate the performance by transferring our \algname{} to different detection pipelines. 
Specially, we simply replace the detection head in Mask RCNN with UniHead to build a mask-based version. 
As is shown in Table~\ref{tab:det_pipelines}, the \algname{} can boost the performance of all these types of detectors, showing its generalization ability on different detection frameworks. 

\begin{table}[t]
\centering
\caption{Results of \algname{} with variant backbones. "DCN" means deformable convolution. * means multi-scale training.}
\scalebox{0.8}{
\begin{tabular}{c| c c c c}
\hline  
Method & Ours & AP & AP$_{.5}$ & AP$_{.75}$\\
\hline
ResNet-50& &  38.8 & 59.9 & 42.1 \\
ResNet-50& \checkmark & \bf{41.8} & \bf{60.6} & \bf{44.7}\\
\hline
ResNet-101& & 39.9 & 60.5 & 43.5 \\
ResNet-101& \checkmark & \bf{42.4} & \bf{61.4} & \bf{45.7} \\
\hline
ResNeXt-101-64x4d& & 42.2 & \bf{63.4} & 45.7 \\
ResNeXt-101-64x4d& \checkmark &\bf{44.5} & 63.2 & \bf{48.0} \\
\hline
ResNeXt-101-64x4d-DCN& & 45.4 & \bf{67.1} & 49.2 \\
ResNeXt-101-64x4d-DCN& \checkmark &\bf{47.3} & 66.9 & \bf{51.3} \\
\hline
Swin-T* & &  43.7& 66.4 & 47.7 \\
Swin-T* & \checkmark & \bf{46.3} & \bf{66.4} & \bf{49.5} \\
\hline
\end{tabular}}
\label{tab:backbones}
\end{table}

\noindent\textbf{Backbone Generalization.}
We further conduct experiments with different backbones under the setting of Faster RCNN. As is shown in Table~\ref{tab:backbones}, our head can steadily boost the performance by $2 \sim 3$\% AP. It demonstrates the generalization ability of our method on variant backbones.

\begin{table}[t]
\centering
\caption{Results on different tasks. "*" indicates multi-scale training, multi-stage refinement and 11x scheduler. "$+$" is multi-scale training and 2x scheduler.}
\scalebox{0.8}{
\begin{tabular}{|c|c|c|c|c c c|}
\hline
Task & Method & backbone & Top-1 acc. & AP & AP$_{.5}$ & AP$_{.75}$ \\ \hline
\multirow{2}{*}{Cls.} & baseline & \multirow{2}{*}{R50} & 78.5 & - & - & - \\ \cline{2-2} \cline{4-7} 
& UniHead &  & \textbf{79.5} & - & - & - \\ 
\hline
\multirow{4}{*}{Det.} & Faster RCNN & \multirow{4}{*}{R50} & - & 38.8  & 59.9 & 42.1  \\  \cline{4-7} 
 & UniHead &  & - &  \textbf{41.8} & \textbf{60.6} & \textbf{44.7} \\ \cline{2-2}
 \cline{4-7} 
 & Mask RCNN&  & - &  39.0 & 59.8 & 42.4 \\ 
 \cline{4-7} 
 & UniHead (box) &  & - &  \textbf{42.3} & \textbf{60.9} & \textbf{45.5} \\ 
\hline
\multirow{2}{*}{Segm.} & DeepSnake~\cite{DeepSnake} & DLA34 & - & 30.3 & - & - \\ \cline{2-7} 
 & UniHead & R50 & - & \textbf{30.4} & 53.2 & 30.1  \\ \hline
\multirow{2}{*}{Keyp.} & PointSet*~\cite{PointSet} & \multirow{2}{*}{R50} & - & \textbf{58.0} & 80.8 & 62.4 \\ \cline{2-2} \cline{4-7} 
 & UniHead$^+$ &  & - & 57.9 & \textbf{82.6} & \textbf{63.9} \\ \hline
\end{tabular}}
\label{tab:tasks}
\end{table}

\noindent\textbf{Task Generalization.}
As mentioned before, our head is a unifying perception head, which means that it can be applied to variant visual tasks. 
To be specific, we use $K=16$ for image classification and object detection, $K=36$ for instance segmentation and $K=17$ points for human pose estimation. 
The baseline of classification is trained with the same setting as \algname{} for fair comparison. The performance is evaluated on ImageNet \textit{val} set for classification, and COCO \textit{val} set for other three tasks. 
The experimental results are shown in Table~\ref{tab:tasks}.
We can see that with a ResNet-50 backbone, the \algname{} makes improvements on classification and object detection, and get a close performance compared with expert models for instance segmentation and pose estimation.

\subsection{Comparison with State-of-the-Art}
We evaluate object detection, instance segmentation and pose estimation on COCO \textit{test-dev}, whose results are shown in Table~\ref{tab:sota_2}. 
The reported AP is related to corresponding tasks, \emph{e.g.}, mask AP for instance segmentation. 
We only adopt multi-scale training for data augmentation and no TTA is used. 
\textbf{It should be noted that we don't introduce any task-aware algorithm design, \emph{e.g.}, multi-stage refinement for pose estimation.}

For object detection, the experimental setting in multi-scale training is $[480,960]$ for image minimum side and $1333$ for image maximum side. We can see that with stronger backbones, our \algname{} can achieve competitive performance, although it is not developed just for object detection. 
For instance segmentation, the same augmentation strategy as object detection is used. Here we also use the mask head of Mask RCNN~\cite{MaskRCNN} to build a mask-based \algname{}. Without bells and whistles, \algname{} gets 46.7\% AP with mask-based head and 39.4\% AP with contour-based head. Compared with expert models, \algname{} achieves comparable performance only using a simpler pipeline.
For pose estimation, we use a larger resolution of input image ($[480,1200]$ for image minimum side and $2000$ for image maximum side). With a surprisingly simple way, \emph{i.e.}, direct keypoint regression using $l_1$ loss, \algname{} gets a close performance compared with other regression-based methods which utilize multi-stage refinement (like \cite{PointSet}) and more iterations of training.

\begin{table}[t]
\centering
\caption{Comparisons of for different algorithms and different tasks evaluated on the COCO \textit{test-dev} set. "FG" and "TG" indicate that the method can be generalized to different visual frameworks and visual tasks, respectively. "*" denotes multi-scale test.}
\begin{center}
\scalebox{0.7}{
\begin{tabular}{c | c c | c | c | c c c | c c c}
\hline  
Method & FG & TG & backbone & iteration &  AP & AP$_{.5}$ & AP$_{.75}$ & AP$_{S}$ & AP$_{M}$ & AP$_{L}$ \\
\hline
\hline
\textbf{Object Detection} && && && && &&   \\
ATSS~\cite{ATSS} & x & x & X-101-64x4d-DCN & 2x & 47.7 & 65.5 & 51.9 & 29.7 & 50.8 & 59.4 \\
BorderDet~\cite{BorderDet} & x & x & X-101-64x4d-DCN & 2x & 48.0 & 67.1 & 52.1 & 29.4 & 50.7 & 60.5 \\
Deformable DETR~\cite{DeformableDETR} & x & x & X-101-64x4d-DCN & $\sim$4x & 50.1 & 69.7 & 54.6 & 30.6 & 52.8 & 64.7 \\
DynamicHead~\cite{DynamicHead} & \checkmark & x & X-101-64x4d-DCN & 2x & 52.3 & 70.7 & 57.2 & 35.1 & 56.2 & 63.4 \\
PointSet~\cite{PointSet} & x & \checkmark & X-101-64x4d-DCN & 2x & 45.1 & 66.1 & 48.9 & - & - & - \\
LSNet~\cite{LSNet} & x & \checkmark & X-101-64x4d-DCN & 2x & 49.6 & 69.0 & 54.1 & 30.3 & 52.8 & 62.8 \\
\hline
\rowcolor{gray!10}
UniHead & \checkmark & \checkmark & X-101-64x4d-DCN & 2x & 50.5 & 70.0 & 54.4 & 31.2 & 53.4 & 64.7 \\
\rowcolor{gray!10}
UniHead & \checkmark & \checkmark & Swin-L & 2x & \textbf{54.7} & \textbf{74.5} & \textbf{59.1} & \textbf{35.6} & \textbf{58.2} & \textbf{70.2} \\
\hline
\hline
\textbf{Instance Segmentation} && && && && &&  \\
\textbf{Mask-based:} && && && && &&  \\
Mask RCNN~\cite{MaskRCNN} & x & \checkmark & X-101-32x4d & 1x & 37.1 & 60.0 & 39.4 & 16.9 & 39.9 & 53.5 \\
HTC~\cite{HTC} & x & \checkmark & X-101-64x4d & $\sim$2x & 41.2 & 63.9 & 44.7 & 22.8 & 43.9 & 54.6 \\
YOLACT~\cite{Yolact} & x & x & ResNet-101 & 4x & 31.2 & 50.6 & 32.8 & 12.1 & 33.3 & 47.1 \\
DetectoRS~\cite{DetectoRS} & x & x & X-101-32x4d & 3x & 45.8 & 69.2 & 50.1 & 27.4 & 48.7 & 59.6 \\
\hline
\rowcolor{gray!10}
UniHead (w/ mask head) & \checkmark & \checkmark & X-101-64x4d-DCN & 3x & 43.6 & 67.1 & 47.0 & 25.1 & 46.5 & 58.1 \\
\rowcolor{gray!10}
UniHead (w/ mask head) & \checkmark & \checkmark & Swin-L & 3x & \textbf{46.7} & \textbf{71.2} & \textbf{50.8} & \textbf{28.2} & \textbf{50.3} & \textbf{62.1} \\
\hline
\textbf{Contour-based:} && && && && &&  \\
ExtremeNet~\cite{ExtremeNet} & x & \checkmark & HG-2 stacked & $\sim$8x & 18.9 & 44.5 & 13.7 & 10.4 & 20.4 & 28.3 \\
DeepSnake~\cite{DeepSnake} & x & x & DLA-34~\cite{DLA} & $~\sim$11x & 30.3 & - & - & - & - & - \\
PolarMask~\cite{PolarMask} & x & x & X-101-64x4d-DCN & 2x & 36.2 & 59.4 & 37.7 & 17.8 & 37.7 & 51.5 \\
PointSet~\cite{PointSet} & x & \checkmark & X-101-64x4d-DCN & 2x & 34.6 & 60.1 & 34.9 & 45.1 & 66.1 & 48.9 \\
LSNet~\cite{LSNet} & x & \checkmark & X-101-64x4d-DCN & $\sim$2x & 37.6 & 64.0 & 38.3 & 22.1 & 39.9 & 49.1 \\
\hline
\rowcolor{gray!10}
UniHead & \checkmark & \checkmark  & X-101-64x4d-DCN & 2x & 36.6 & 63.0 & 36.2 & 22.0 & 38.6 & 48.5 \\
\rowcolor{gray!10}
UniHead & \checkmark & \checkmark & Swin-L & 2x & \textbf{39.4} & \textbf{67.0} & \textbf{39.3} & \textbf{24.7} & \textbf{41.7} & \textbf{52.0} \\
\hline
\hline
\textbf{Pose Estimation} && && && && &&  \\
\textbf{Heatmap-based:} && && && && && \\
CPN~\cite{CPN} & x & x &ResNet-Inception & - & 72.1 & 91.4 & 80.0 & - & 68.7 & 77.2 \\
HRNet~\cite{HRNet} & x & x & HRNet-W48 & $\sim$16x & 75.5 & 92.5 & 83.3 & - & 71.9 & 81.5 \\
DARK ~\cite{DARK} & x & x & HRNet-W48 & $\sim$11x & \textbf{76.2} & \textbf{92.5} & \textbf{83.6} & - & \textbf{72.5} & \textbf{82.4} \\
\hline
\textbf{Regression-based:} && && && && &&  \\
CenterNet*~\cite{CenterNetObj} & x & \checkmark & HG-2 stacked & $\sim$11x & 63.0 & 86.8 & 69.6 & - & 58.9 & 70.4 \\
PointSet~\cite{PointSet} & x & \checkmark & X-101-64x4d-DCN & $\sim$8x & 62.5 & 83.1 & 68.3 & - & - & - \\
LSNet~\cite{LSNet} & x & \checkmark & X-101-64x4d-DCN & $\sim$6x & 59.0 & 83.6 & 65.2 & - & 53.3 & 67.9 \\
\hline
\rowcolor{gray!10}
UniHead & \checkmark & \checkmark & X-101-64x4d-DCN & 2x & 65.4 & 87.3 & 72.6 & - & 60.9 & 72.3  \\
\rowcolor{gray!10}
UniHead & \checkmark & \checkmark & Swin-L & 2x & \textbf{66.1} & \textbf{88.7} & \textbf{73.7} & - & \textbf{62.0} & \textbf{72.3} \\
\hline
\end{tabular}}
\end{center}
\label{tab:sota_2}
\end{table}

\section{Conclusion}
In this paper, we proposed \algname{}, a unifying visual perception head. It can not only be embedded in variant detection frameworks, but also applied to different visual tasks, including image classification, object detection, instance segmentation and pose estimation. \algname{} perceives instances by dispersible points learning, which is also equipped with transformer encoders to capture semantic relations of them. Though our \algname{} is designed in a simple way, it achieves comparable performance on each task compared with expert models. This work shows the potential in general visual learning and we hope it can promote universal visual perception research.

\noindent\textbf{Acknowledgement:}
The work was supported by the National Key R\&D Program of China under Grant 2019YFB2102400.

\clearpage
%
%
\bibliographystyle{splncs04}
\bibliography{egbib}

\clearpage
\appendix

\section{Detailed Architecture of UniHead}
\begin{figure}[h]
  \centering
  \includegraphics[width=1.0\linewidth]{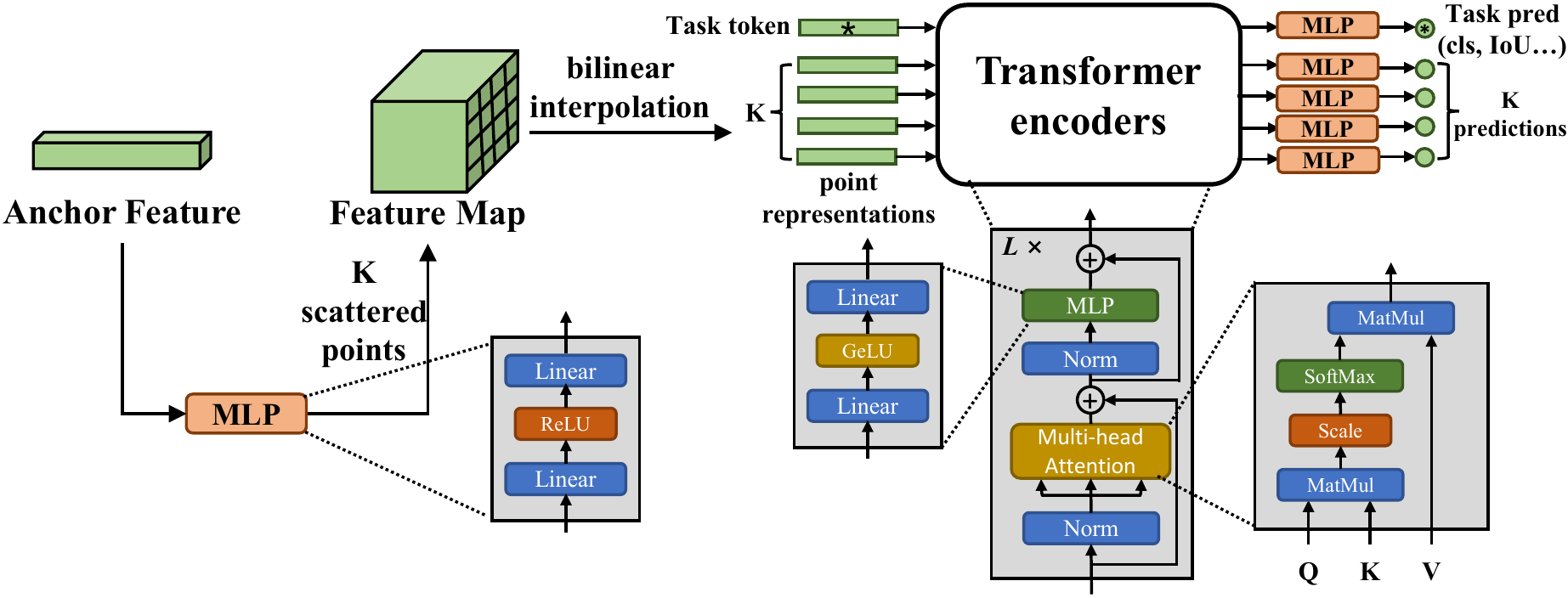}
  \caption{An illustration of \algname{} detailed architecture.}
  \label{fig:detail}
\end{figure}
A detailed illustration is given in Fig.\ref{fig:detail}. First, K scattered points are generated through an multi-layer perceptron and we use bilinear interpolation to extract point representations. Then, together with a learnable embedding vector (optional), they are fed into transformer encoders. Finally, we use different MLPs to get the final results, where ``task pred" means predictions from the task token, like IoU token.

\section{Discussions on UniHead Unification}
The unifying concept in \algname{} indicates that it can be easily transferred to various visual tasks with different visual frameworks without complex adaptations and adjustments. 
In Table~\ref{tab:unify_comp}, we also show some methods with generalization abilities. 
They use point-based~\cite{CenterNetObj,LSNet,PointSet} or mask-based localization \cite{MaskRCNN} to support multiple tasks, or achieve framework generalization by unifying head design~\cite{DynamicHead}.
But it is hard to perform various visual tasks with different visual frameworks in a single method.

In addition, the differences with similar methods are also discussed.

\noindent\textbf{Mask R-CNN.}
Mask R-CNN introduces a strong baseline for mask-based segmentation methods. 
The mask prediction is made on the upsampled feature map to achieve higher performance. 
Inevitably, this process introduces extra computational overhead.
The dependence on proposal features also makes it hard to be transferred to the one-stage framework.
\algname{} adopts point-based segmentation methods, which greatly decrease the computational cost by predicting the sparse points located in the contour of an instance.
For example, Mask R-CNN costs \textbf{68 GFLOPs} for mask generation while \algname{} only needs \textbf{5 GFLOPs}.
But the modeling error for more complex objects is still challenging in this research community.
Besides, in our design, an anchor point is all \algname{} needs to generate a mask, so it can be easily adopted to the one-stage framework.
We aim to conduct a simple but general \algname{}
for different frameworks and visual tasks instead of addressing these task-specific bottlenecks.


\noindent\textbf{PointSet and LSNet.}
PointSet~\cite{PointSet} and LSNet~\cite{LSNet} both use different number of points to adapt to different tasks. Following the one-stage detection process, they make dense predictions on different positions of feature maps, which is similar to the FCOS~\cite{FCOS} architecture. 
\algname{} is not only applicable to the one-stage detectors, but also to the situation where there exists regions of interest, \emph{i.e.}, Faster-RCNN-like frameworks. 
We also propose dispersible points learning to effectively extract decision-relevant token features, which can automatically adapt to different visual task requirements. Furthermore, we perform token-to-token comparison to enhance global information by transformer encoders. 
It makes \algname{} more convenient to transfer to different tasks and architectures directly.

\begin{table}[t]
\centering
\caption{Comparison with other methods from two perspectives: task generalization and framework generalization.}
\begin{tabular}{|c|cccc|cc|}
\hline
\multirow{2}{*}{Method} & \multicolumn{4}{c|}{Task} & \multicolumn{2}{c|}{Framework} \\ \cline{2-7} 
 & \multicolumn{1}{c|}{Cls.} & \multicolumn{1}{c|}{Det.} & \multicolumn{1}{c|}{Segm.} & Keyp. & \multicolumn{1}{c|}{One-stage} & Two-stage \\ \hline
Mask R-CNN\cite{MaskRCNN} & \multicolumn{1}{c|}{} & \multicolumn{1}{c|}{\checkmark} & \multicolumn{1}{c|}{\checkmark} & \checkmark & \multicolumn{1}{c|}{} & \checkmark \\ \hline
Dynamic Head\cite{DynamicHead} & \multicolumn{1}{c|}{} & \multicolumn{1}{c|}{\checkmark} & \multicolumn{1}{c|}{} &  & \multicolumn{1}{c|}{\checkmark} & \checkmark  \\ \hline
CenterNet\cite{CenterNetObj} & \multicolumn{1}{c|}{} & \multicolumn{1}{c|}{\checkmark} & \multicolumn{1}{c|}{} & \checkmark  & \multicolumn{1}{c|}{\checkmark} &  \\ \hline
PointSet\cite{PointSet} & \multicolumn{1}{c|}{} & \multicolumn{1}{c|}{\checkmark} & \multicolumn{1}{c|}{\checkmark} & \checkmark  & \multicolumn{1}{c|}{\checkmark} &  \\ \hline
LSNet\cite{LSNet} & \multicolumn{1}{c|}{} & \multicolumn{1}{c|}{\checkmark} & \multicolumn{1}{c|}{\checkmark} & \checkmark  & \multicolumn{1}{c|}{\checkmark} &  \\ \hline
UniHead & \multicolumn{1}{c|}{\checkmark} & \multicolumn{1}{c|}{\checkmark} & \multicolumn{1}{c|}{\checkmark} & \checkmark  & \multicolumn{1}{c|}{\checkmark} & \checkmark  \\ \hline
\end{tabular}
\label{tab:unify_comp}
\end{table}

\section{Discussion on Dispersible Points}
\begin{figure}[t]
  \centering
  \includegraphics[width=0.6\linewidth]{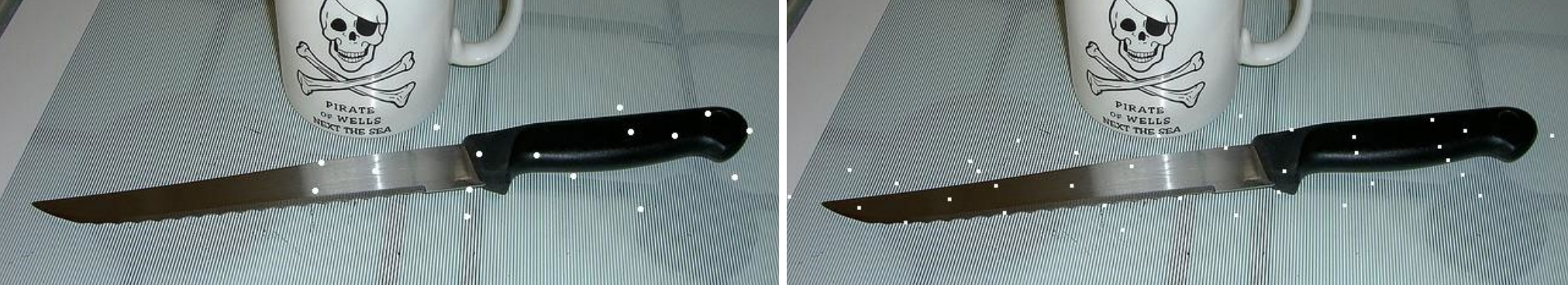}
  \caption{Visualization of dispersible points for classification (left) and segmentation (right).}
  \label{fig:cls_seg}
\end{figure}
Different tasks require different feature representations.
As illustrated in Fig.\ref{fig:cls_seg}, dispersible points for classification focus more on the salient part of an instance like knife handles, while for segmentation they are located at the contour of the instance.
In \algname{}, the learning of dispersible points is not supervised in a direct way, so that the extracted features can be decision-relevant and automatically adapt to different visual task requirements.

\section{Qualitative Results}
\begin{figure}[h]
  \centering
  \includegraphics[width=1.0\linewidth]{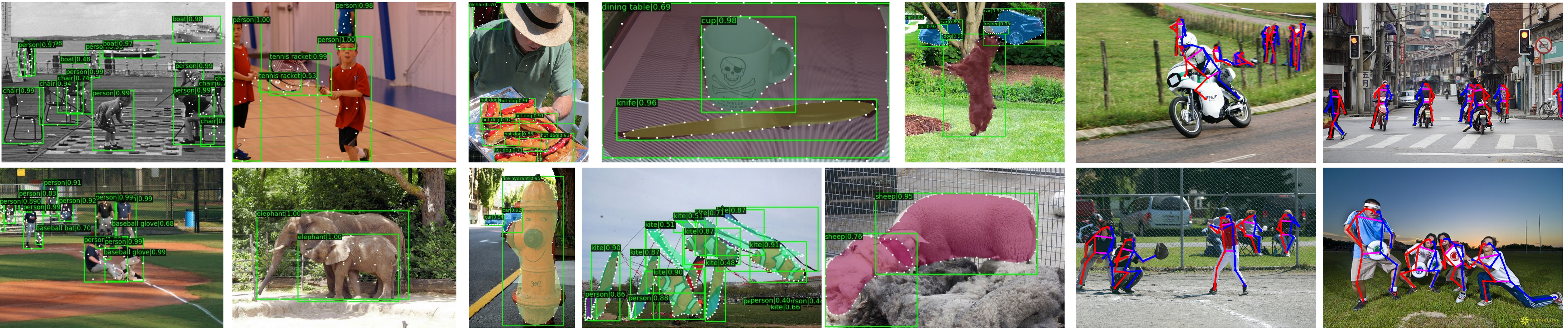}
  \caption{A visualization of \algname{} for object detection, instance segmentation and pose estimation on COCO \textit{val} set. Localization points for detection/segmentation are viewed in white dots. Only results with scores higher than 0.4 are shown.}
  \label{fig:vis}
\end{figure}

We show some visualization results of \algname{} on object detection, instance segmentation, and pose estimation. See Fig.\ref{fig:vis}.

\end{document}